# Layer-wise Regularized Adversarial Training

# using Layers Sustainability Analysis framework

**Mohammad Khalooei[1], Mohammad Mehdi Homayounpour[2], Maryam Amirmazlaghani[3]**

[1] khalooei@aut.ac.ir, [2] homayoun@aut.ac.ir, [3] mazlaghani@aut.ac.ir

[1,2,3]Department of Computer Engineering and Information Technology, Amirkabir University of

Technology, No. 350, Hafez Ave, Valiasr Square, Tehran, Iran

**Abstract**

Deep neural network models are used today in various applications of artificial intelligence, the strengthening of which, in the face of adversarial attacks is of particular importance. An appropriate solution to adversarial attacks is adversarial training, which reaches a trade-off between robustness and generalization. This paper introduces a novel framework (Layer Sustainability Analysis (LSA)) for the analysis of layer vulnerability in an arbitrary neural network in the scenario of adversarial attacks. LSA can be a helpful toolkit to assess deep neural networks and to extend the adversarial training approaches towards improving the sustainability of model layers via layer monitoring and analysis. The LSA framework identifies a list of Most Vulnerable Layers (MVL list) of the given network. The relative error, as a comparison measure, is used to evaluate representation sustainability of each layer against adversarial inputs. The proposed approach for obtaining robust neural networks to fend off adversarial attacks is based on a layer-wise regularization (LR) over LSA proposal(s) for adversarial training (AT); i.e. the AT-LR procedure. AT-LR could be used with any benchmark adversarial attack to reduce the vulnerability of network layers and to improve conventional adversarial training approaches. The proposed idea performs well theoretically and experimentally for state-of-the-art multilayer perceptron and convolutional neural network architectures. Compared with the AT-LR and its corresponding base adversarial training, the classification accuracy of more significant perturbations increased by 16.35%, 21.79%, and 10.730% on Moon, MNIST, and CIFAR-10 benchmark datasets, respectively. The LSA framework is available and published at https://github.com/khalooei/LSA.

# 1  Introduction

Deep Neural Networks (DNNs) have thus far achieved great success in various fields and tasks of artificial intelligence, including computer vision, speech processing, natural language processing, and time-series analysis [1–4]. With the popularity of DNN tools in different tasks of human life, they are required to pass certain standardization milestones. It has been demonstrated that small targeted changes in DNNs' inputs, called perturbations, can easily fool DNNs, exposing them to vulnerabilities. It should be noted that these slight changes are different from usual statistical noises and are near the worst-case scenario for perturbations in adversarial cases [5,6]. Schedey *et al.* [5] called these perturbed inputs *Adversarial examples*. Goodfellow *et al.*[6] further elaborated on the concept of adversarial example, bringing theoretical explanation alongside experimental analysis to substantiate the presence of adversarial examples. Several other works have also divulged vulnerabilities of DNNs in different conditions; a White-Box for example is when the attacker has completed access to the model parameters and architecture, while a Black-Box is one where the attacker does not have access to the model parameters or architecture [7]; there are adversarial attacks in which the adversarial example is obtained in a single step using a gradient signal, while others use multi-step guidelines for creating adversarial examples [8]. An adversarial example is obtained by limiting the amount of perturbation or defining the parameters of the transformation attack [9].

In response to such adversarial attacks, measures have been taken to design high-adequacy classifiers for different attack scenarios [6,8,10–16]. Empirical risk minimization, in this line, has been more successful at finding a classifier model with a small error, but it does not provide model robustness [17]. There is currently a challenge in finding a proper method for analyzing the behavior of a neural network and determining the most effective defense approach against adversarial attacks and state-of-the-art attacks. Exercising on the transformation, refinement, and reconstruction of inputs would be a way to attain greater robustness [13]. Likewise, certain adversarial example detection strategies concentrate on disentangling clean and malicious inputs as a pre-active method [18,19]. Optimization-based approaches intuitively train the network by optimizing the network parameters to resist the worst-case example inside the predefined boundary [8,20]. Adversarial training (AT) is essentially a method of increasing robustness by modeling and solving an optimization problem.

Using an ensemble of popular adversarial attack approaches is one way of enhancing the robustness of adversarially trained models [21]. Baseline adversarial training approaches exploit the knowledge embedded in the adversarially perturbed inputs [22]. However, most of the attention in the mentioned works is on the loss function, apart from recent researches on the middle layers as complement to the loss function [23–25]. In contrast to adversarial model robustness, layer-wise ideas aim to move perturbation findings from the input layer into the middle layer in order to enhance the generic model's robustness. However, they need to change the architecture and learning process to accommodate their approach, which is more time-consuming.

This work proposes a framework for assessing the representation outputs of neural network layers called Layer Sustainability Analysis (LSA), which attempts to evaluate neural network vulnerability behaviors in order to reduce layer vulnerability based on adversarial training guidelines. The LSA framework procures a list of Most Vulnerable Layers called the LSA MVL list or LSA proposals. Each LSA proposal can participate in an Adversarial Training (AT)-based procedure named Layer-wise Regularized (LR) or AT-LR procedure. LSA improves our explanatory competence of the sustainability of each layer in the network against input perturbations. Using AT-LR along with related ideas makes the model more stable and robust compared to the standard one. Furthermore, experiments on the benchmark datasets confirm that the proposed technique could be a good solution for layer-wise analysis of DNNs and improve adversarial training approaches. The main contributions of this paper can thus be summarized along these lines:

- The *layer sustainability analysis* (LSA) framework is introduced to evaluate the behavior of layer-level representations of DNNs in dealing with network input perturbations using Lipschitz theoretical concepts.
- A layer-wise regularized adversarial training (AT-LR) approach significantly improves the generalization and robustness of different deep neural network architectures for significant perturbations while reducing layer-level vulnerabilities.
- AT-LR loss landscapes for each LSA MVL proposal can interpret layer importance for different layers, which is an intriguing aspect

The organization of this paper is as follows; Section 2 reviews preliminaries and related works. Section 3 and 4 describe the proposed method and experimental results; and finally, Section 5 presents conclusions and future works.

## 2  Related Works

This section briefly covers the main ideas in recent adversarial training based approaches. In terms of learning a robust model, many efforts come to deal with the problem of adversarial examples. And so, the following will first present a definition of adversarial example and adversarial attack then go further into the concepts of adversarial defense and adversarial training and their extensions as efficient defense measures.

Simply explained, any perturbed sample that fools neural network models using imperceptible perturbations for humans are considered adversarial examples. To obtain adversarial examples, Goodfellow *et al.* [6] proposed an optimization problem by introducing a straightforward gradient-based process called the *Fast Gradient Sign Method* (FGSM); where if $x$, $\hat{x}$ and $\varepsilon$ represent input sample, corresponding adversarial example, and perturbation rate for a specific classifier model $F$, a simple constraint $\|\hat{x} - x\|_\infty \leq \varepsilon$ can be used to restrict the adversarial example $\hat{x}$ over $L_\infty$ ball and maximize the loss function $J(\theta, \hat{x}, y)$ to fool DNNs through FGSM pipeline as formulated in equation (1):

$$\hat{x} = x + \varepsilon \cdot sign\{\nabla_x J(\theta, x, y)\}, \tag{1}$$

where $sign(\cdot)$ and $\nabla_x J(\theta, x, y)$ denote the signum function and the gradient of the loss function of classification model $F$ with parameters $\theta$ for input $x$ and its corresponding label $y$. Kurakin *et al.* [26] introduced the iterative version of FGSM to obtain a proper adversarial example in a multi-step method. By oscilating and moving in the infinity norm bound on all dimensions along with the gradient ascent direction, each data point which fools the classifier is considered an adversarial example. Madry *et al.* [8] provided a multi-step approach called *Projected Gradient Descent* (PGD), where it proceeded several steps to obtain the adversarial example. The main challenge in this regard is to solve the optimization problem in equation (2) and find the best malicious or adversarial samples $\hat{x}$.

$$\hat{x} = \max_{x^* \in B(X, \varepsilon)} J(\theta, x^*, y), \tag{2}$$

The $B$ function refers to the bounded space where the attacker restricts its diversity and may relate to parameters like perturbation rate $\varepsilon$ as mentioned in [6] or transformation attack parameters as discussed in [9]. Croce and Hein [27] on the other hand have concentrated on the steps of the PGD attack Auto-PGD (APGD). They partitioned whole steps into two phases; an exploration phase finds a feasible set for an initial proper sample point, and an exploitation phase provides an effective adversarial example. This partitioning progressively and automatically reduces the step size in an optimization process. When the rise in the significance of the objective is sufficiently fast, the step size is deemed reasonable; otherwise, the size needs to be reduced. To demonstrate the existence of adversarial examples, authors of [12,28,29] interpreted that they are likely due to computational hardness.

Adversarial defense techniques primarily attempt to hold off different gradient-based or non-gradient-based accesses (styles of attack). Some of these concentrate on the training phase as a pre-active solution [6,8,37–40,14,30–36], and others work in the inference phase as a proactive solution [17,41,50,51,42–49]. Others like Fawzi *et al.* [52]

have pursued robustness bounds concentration using the smooth generative model to place an upper bound on robustness.

Adversarial training (AT) is the name indicated for constituting robustness in the setting of an optimization problem, formulated in equation (3):

$$\min \mathbb{E}_{(x,y)\sim \mathbb{D}} \left\{ \max_{\hat{x}\in B(x,\varepsilon)} J(\theta, \hat{x}, y) \right\}, \tag{3}$$

Goodfellow *et al.* [6] indicated that training with an attack-specific adversarial loss function, such as the FGSM, could affect the primary loss function and improve its robustness. They denoted their extension as equation (4):

$$\tilde{J}(\theta, x, y) = \delta J(\theta, x, y) + (1 - \delta) J(\theta, x + \varepsilon \cdot sign(\nabla_x J(\theta, x, y)), y), \tag{4}$$

where $\tilde{J}$ is the adversarial training objective function belonging to the FGSM attack, and $\delta$ is a parameter of a linear combination for each regularization term. Madry *et al.* [8] rephrase equation (3) with the definition of population risk by incorporating a projected gradient-based adversary titled PGD adversarial training, correspondent with the proposed AT-PGD terminology in this paper. They proposed a saddle point problem (equation (5)) as the composition of an inner maximization and an outer minimization problem.

$$\min_{\theta} \rho(\theta) \quad s.t. \rho(\theta) = \mathbb{E}_{(x,y)\sim \mathbb{D}} \left\{ \max_{\hat{x}\in B(x,\varepsilon)} J(\theta, \hat{x}, y) \right\}, \tag{5}$$

where $J$, $\theta$ and $\hat{x}$ are in respective order, the objective function of the problem, the model parameters, and the adversarial examples (corresponding to input sample $x$). Each adversarial example $\hat{x}$ is restricted by a bound $B$ with radius $\varepsilon$. The inner attack problem is set to find a high objective function value for determining an adversarial edition of the given data point. In contrast, the outer minimization problem seeks to minimize the adversarial loss rising from the inner attack problem. Wong and Kolter [53] supplied a guarantee over a deep ReLU based classifier against any norm-bounded variation of inputs. They provided an adversarial polytope as a convex outer bound for a set of last layer activation maps, introducing a provable robust deep classifier which works by computing a feasible dual optimization problem solution. Other works have also considered a lagrangian penalty formulation for their optimization problem [27,37,54–56].

Adversarial training effectively robustifies models but decreases accuracies over clean samples [57]. It also suffers from the problem of overfitting to adversarial samples used in training procedures, as discussed in more detail in [58–60]. A theoretical principled trade-off was introduced by Zhang *et al.* [32] between robustness and accuracy; i.e. the TRADE adversarial training approach (TRadeoff-inspired Adversarial DEfense via Surrogate-loss minimization).

Their approach worked through pushing the decision boundary of the classifier away from the sample by minimizing the comparison measure between the prediction values For clean sample $f(x)$ and adversarial example $f(\hat{x})$ as shown below:

$$\min_{f} \mathbb{E}_{(x,y)\sim\mathbb{D}} \left\{ L(f(x), y) + \lambda \max_{\hat{x} \in B(x,\varepsilon)} L(f(x), f(\hat{x})) \right\}, \tag{6}$$

where $\lambda$ is a coefficient indicating the rigidity and strength of regularization and plays a crucial role in balancing the importance of clean and robust errors. Also, the loss function $L$ in the first and second terms of the objective function of equation (6) indicate cross-entropy and classification-calibrated loss, respectively. As denoted in [32], the TRADE adversarial training (AT-TRADE) method surpasses PGD adversarial training in terms of accuracy. Wong *et al.* [61] demonstrated that AT-FGSM could fail due to catastrophic overfitting and introduced a similar approach for random initialization in FGSM attack. Their approach is denoted as a fast adversarial training (AT-FAST) method that is as efficient as AT-PGD. This being so, Andriushchenko and Flammarion [62] identified certain shortcomings of AT-FAST in catastrophic overfitting and zero-initialized perturbations.

Adversarial training on the middle layer was also proposed from a different point of view by Sabour *et al.* [23], who showed that representations of each layer in DNNs can slightly manipulate and change the classifier's decision. They focused on the internal layer of DNN representations to obtain a novel type of adversarial examples that differs from other conventional adversarial attacks, as shown in equation (7).

$$\hat{x} = arg \min_{x} \|\phi_k(x) - \phi_k(x_g)\|_2^2 \quad s.t. \quad \|x - x_s\|_\infty < \varepsilon, \tag{7}$$

where $\phi_k$ and $x_s$ are the representation of layer k and source input sample, respectively and, $x_g$ denotes a target or guide sample. In addition, $\hat{x}$ is a close sample to the source sample $x_s$. The constraint on the distance between $\hat{x}$ and $x_s$ is formulated in terms of the $l_\infty$ norm to restrict adversarial example $\hat{x}$ to the $\varepsilon$-based bound. Chen and Zhang in [25] proposed a layer-wise approach and concentrated on layer adversarial perturbations acquainted in middle layers. They implemented layer-wise adversarial training for all layers, which is much time-consuming due to adversarial training for each layer through the training process. Sankaranarayanan *et al.* [24] attempted to regularize DNN by perturbing middle layer activations. They observed that adversarial perturbations generalize across different samples for hidden layer activations. Their observation devises an efficient regularization approach to help the training of very deep architectures. Albeit, the proposed method was inferior to dropout generalization but succeeded in enhancing adversarial robustness. These extensions of adversarial training are still in progress, to acheieve as much robustness as possible. It is worth noting that the analytical point of view for choosing layers in layer-wise approaches is essential. With this background, the following sections will introduce theoretical and practical approaches in analyzing the neural network model and choosing a critical layer for the adversarial training approach.

## 3   Proposed method

This section explains the proposed LSA framework and highlights Layer-wise Regularized (LR) adversarial training contributions over LSA proposals. First, a layer-wise sustainability analysis framework is described. Then, an LR adversarial training (AT-LR) methodology is presented to deal with the vulnerabilities of neural network layers.

### 3.1   Layer Sustainability Analysis (LSA) framework

Sustainability and vulnerability in different domains have many definitions. In our case, the focus is on certain vulnerabilities that fool deep learning models in the feed-forward propagation approach. Our main concentration is therefore on the analysis of forwarding vulnerability effects of deep neural networks in the adversarial domain. Analyzing the vulnerabilities of deep neural networks helps better understand different behaviors in dealing with input perturbations in order to attain more robust and sustainable models. One of the fundamental mathematical concepts that comes to mind in the sustainability analysis approach is Lipchitz continuity which grants deeper insight into the sustainability analysis of neural network models by approaching LR from the Lipschitz continuity perspective. Let $F$ be a function that satisfies the Lipschitz condition in the variable $X$. For any such $F$, assume a constant $\psi > 0$ as the smallest number which satisfies the following inequality:

$$\|F(x_1) - F(x_2)\| \leq \psi \|x_1 - x_2\| \qquad s.t.\ x_1, x_2 \subset X. \tag{8}$$

The smallest $\psi$ is the best Lipschitz constant. It means that, when two inputs $x_1$ and $x_2$ differ slightly, the difference between their corresponding outputs of $F$ is also small. The functionality of inequality (8) can be applied to any layer in a neural network. For this, we may replace $F$ by $\phi_l$ where $\phi_l(x_1)$ and $\phi_l(x_2)$ are the output representation tensor of layer $l$ for network input samples $x_1$ and $x_2$, respectively. Therefore, inequality (8) can be redefined as

$$\|\phi_l(x_1) - \phi_l(x_2)\| \leq \psi \|x_1 - x_2\|. \tag{9}$$

If $x_1$ is a clean sample $x$ and $x_2$ is its corresponding adversarial example $\hat{x}$, inequality (9) can be rewritten as

$$\|\phi_l(x) - \phi_l(\hat{x})\| \leq \psi \|x - \hat{x}\|. \tag{10}$$

**Algorithm 1 .** Algorithm to find the most vulnerable layers in the layer sustainability analysis (LSA) framework

**Input**: Output representation tensors $\phi_l(x)$ and $\phi_l(\hat{x})$ of layer $l$ for clean input $x$ and the corresponding perturbed sample $\hat{x}$

**Output**: list of most vulnerable layers

**Algorithm steps:**

for trained model $m$, constant $\eta$, average $\mu$ and standard deviation $\sigma$ as calculated in equation (12).

1. LSA_MVL_list = []
2. for $l$ in range(0, $Ly$)
3.     if$\left(CM\big(\phi_l(x), \phi_l(\hat{x})\big) - \mu\right) > \eta\, \sigma$
4.         LSA_MVL_list.append($l$)
5. LSA_MVL_list = sort(LSA_MVL_list)

Inequality (10) implies that if two similar inputs (clean and its corresponding perturbed one) are fed to the neural network, the representation tensor $\phi_l(\cdot)$ of each layer of neural network corresponding to clean and perturbed inputs must also be similar. The following equations are based on inequality (10) to determine the vulnerability effects of the neural network layers and prepare some indicators to identify vulnerable layers and analyze the response of the layers.

$$CM\big(\phi_l(x), \phi_l(\hat{x})\big) = \frac{\|\phi_l(x) - \phi_l(\hat{x})\|_F}{\|\phi_l(x)\|_F}, \tag{11}$$

where $CM$ denotes the comparison measure as a relative error between two representation tensors $\phi_l(x)$ and $\phi_l(\hat{x})$ of layer $l$ in the network architecture. The LSA framework then assesses the output representation tensor of each layer and to distinguish the vulnerable layer, parameters average $\mu$ and standard deviation $\sigma$ are calculated using the comparison measure of each layer as shown in equation (12), and are used in Algorithm 1 to find the most vulnerable layers.

$$\mu = \frac{1}{M \times Ly} \sum_{m=0}^{M-1} \sum_{l=0}^{Ly-1} CM\big(\phi_l(x_m), \phi_l(\hat{x}_m)\big),$$

$$\sigma = \sqrt{\frac{1}{M \times Ly} \sum_{m=0}^{M-1} \sum_{l=0}^{Ly-1} \big(CM\big(\phi_l(x_m), \phi_l(\hat{x}_m)\big) - \mu\big)^2}, \tag{12}$$

where $Ly$ is the number of learnable layers—learnable layer meaning any layer which has weights, such as learnable convolutional or fully connected layers. Moreover, $M$ is the number of randomly selected samples in the training set and is chosen to be much less than the total number of training samples to decrease the computation time. In Algorithm 1, the combined parameter $\eta$ represents cut-off threshold for parameters that are set for the detection of vulnerability, which is crucial to determining vulnerable layers in different networks.

After obtaining the most vulnerable layer proposals (MVL list), each proper MVL proposal is selected to be used in the LR adversarial training (AT-LR). One primary strength of the proposed approach is its ability to reduce the vulnerability of layers. Finding more vulnerable layers has a significant impact on the durability of layers and the generalization and robustness of the network. To demonstrate this, Section 4 conducts several experiments on different model architectures and scenarios in order to consolidate the following Algorithm 2 for the proposed layer sustainability analysis (LSA) framework.

Algorithm 2. Layer sustainability analysis (LSA) framework's algorithm

---

**Algorithm 2** Layer Sustainability Analysis (LSA) framework

**Input**: model $m$, train data $D_{train}$, test data $D_{test}$, attack method and its parameters

**Output**: list of vulnerable layers

**Algorithm steps**:

1. Standard training of model $m$ using $D_{train}$ samples
2. Execute attack on the trained model $m$ and obtain adversarial examples by perturbing $D_{test}$ samples
3. Run Algorithm 1 and tune the proper cut-off threshold $\eta$ of Algorithm 1 to find out the MVL list
4. Return the MVL list of Algorithm 1

---

Fig. 1 illustrates the LSA algorithm diagram for layerwise analysis of neural networks. As depicted in Fig. 1, the clean and perturbed input data pass through the trained network in the LSA framework. LSA assesses the output representation tensors of each layer corresponding to the network clean and perturbed input samples using Algorithm 1. Layer sustainability analysis for clean and corresponding adversarial input can detect traces of vulnerability through the network layers.

VGG [63] network architecture is used in Fig. 2 only as an illustration of the different layers' behavior and the corresponding LSA measure values. Fig. 2 illustrates the LSA results for all network layers and shows that the comparison measure value fluctuation curves for different statistical and adversarial perturbations are similar.

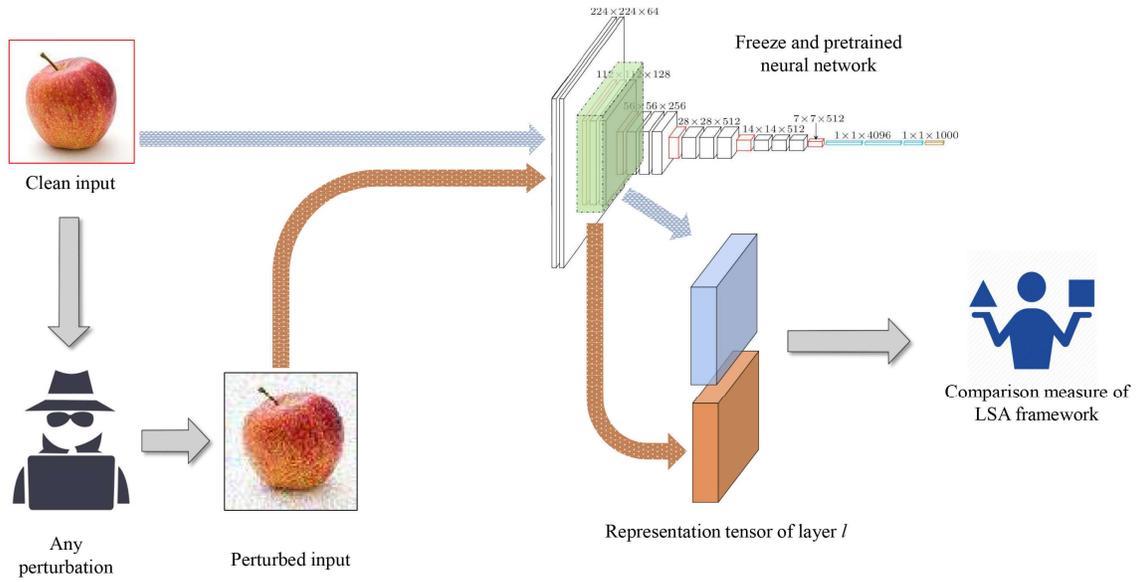

**Fig. 1.** Diagram of the Layer Sustainability Analysis (LSA) framework

Note that some layers are more vulnerable than others. In other words, some layers are able to sustain disruptions and focus on vital features, while others are not. Each layer in Fig. 2 is related to any of learnable convolutional or fully connected layers.

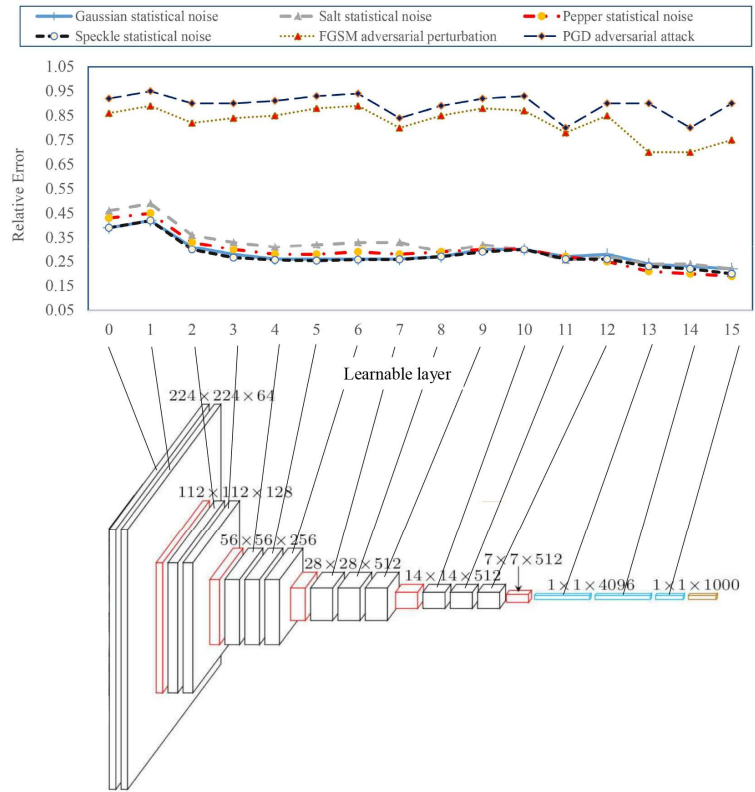

**Fig. 2.** Comparison measure values for corresponding layers of the VGG network in the proposed LSA framework

The next section explains the proposed methodology for controlling vulnerabilities by inducing a regularization term defined in the neural network optimization phase of the loss function.

### 3.1.1 Layer-wise Regularized Adversarial Training (AT-LR)

One of the incentives of introducing regularization terms in the loss function of deep neural networks is to restrict certain effective parameters. Researchers have attempted to discover effective parameters in several ways, but most approaches are not applicable to all networks. This paper presents a new approach to perform an effective sensitivity analysis of different middle layers of a neural network and administer the vulnerability in the loss function. The loss function of the network can be improved by including such regularization terms to reduce the vulnerability of middle layers. As observed in equations (13) and (14), the proposed LR term is added in order to define an extension on base adversarial training through an inner maximization and outer minimization optimization problem.

$$\hat{x} = \arg\max_{Z \in B(x,\varepsilon)} J(\theta, Z, y), \tag{13}$$

$$\min \mathbb{E}_{(x,y)\sim\mathbb{D}}\{J(\theta, \hat{x}, y) + LR(\theta, x, \hat{x}, y)\}, \tag{14}$$

where $B(x, \varepsilon)$ is a ball around sample point $x$ with $\varepsilon$ as its radius. Moreover, LR uses base network weights $\theta$, input sample $x$ and corresponding label $y$ and its related adversarial example $\hat{x}$—which is prepared in an adversarial attack approach. Although two different adversarial attacks can be used for the first and the second terms of the minimization problem (14), however for simplicity, the same adversarial attack obtained from equation (13), i.e. $\hat{x}$, is utilized. Also, any benchmark adversarial training approach can be embedded in the LR adversarial training (AT-LR). To introduce our AT-LR loss, related comparison measure assessment steps are used as mentioned in the LSA framework. As discussed before, in Algorithm 2, the comparison measure is calculated for each of the most vulnerable layers listed in LSA MVL list. Equation (15) is defined as a combined regularization term based on equation 13 to deal with the vulnerability of each vulnerable LSA proposal:

$$LR(\theta, x, \hat{x}, y) = \sum_{l \in \mathcal{M}} \gamma_l CM\big(\phi_l(x), \phi_l(\hat{x})\big), \tag{15}$$

where $\mathcal{M}$ is the MVL list obtained from Algorithm 2, and $\gamma_l$ is the constant parameter for each layer $l$. Algorithm 3 determines the outline of the AT-LR adversarial training procedure. As mentioned earlier, an adversarial example $\hat{x}$ is provided from any adversarial attack approach (such as FGSM, PGD, FAST, APGD, etc.). The critical point of this approach is based on the main steps of LSA in Algorithm 2. With this idea, differences in representation tensors of layers are reduced while the clean sample and its perturbed version with a slight variation are fed into the neural network, inducing reductions in the Lipschitz constant. The process also improves generalization and robustness simultaneously.

---

**Algorithm 3** Layer-wise Regularized adversarial training (AT-LR) algorithm

---

**Input**: X as inputs, Y as the corresponding targets, $F_\theta$ as a model with parameters $\theta$, an LSA MVL list from Algorithm 2

**Output**: a robust model (based on AT-LR approach)

**Algorithm steps:**

1. Initialize $\theta$
2. **for** epoch =1 … N **do**
3.     **for** minibatch $(x, y) \subset (X, Y)$ **do**
4.         $\hat{x} \leftarrow \text{AdversarialAttack}(F_\theta, x, y)$
5.         $\theta \leftarrow \min \{ J(\theta, \hat{x}, y) + LR(\theta, x, \hat{x}, y) \}$
6.     **end for**
7. **end for**

---

The next section covers experiments to demonstrate the aforementioned proposed ideas.

# 4 Evaluation

This section provides an experimental analytical perspective on the proposed LSA framework and the LR adversarial training approach named AT-LR. The acronym AT-FGSM-LR denotes regularized adversarial training with the FGSM approach. In the following experiments, we cover different adversarial training approaches like AT-PGD [8], AT-TRADE [32], AT-FAST [61], and AT-APGD [27]. Experimental setup configurations are also discussed, with further reports and ablation studies on experiments and their analysis.

## 4.1 Experimental setup

A diverse set of configurations were used in the following experiments to test different analytical perspectives. The benchmark datasets, network architectures, perturbations, and scenarios used in the experiments are described below.

### 4.1.1 Dataset description

To determine the effectiveness and feasibility of the proposed framework, three widely used datasets, including Moon [64], MNIST [65], and CIFAR-10 [66], which are commonly used to examine the performance of different approaches, were used. The moon dataset consists of 2D features that visualize two interleaving half-circles generated by the Scikit-learn popular python library [67]. A view of the Moon dataset is illustrated in Fig. 3 (a). The MNIST dataset comprises 28×28 black and white images (one channel) of handwritten digits. It contains 60000 and 10000 training and test samples, respectively (Fig. 3 (b)). The MNIST classes are denoted as a label of each handwritten digit from zero to nine (10 categories). The CIFAR-10 consists of 32×32 color images (three channel), with 60000 and 10000 samples for training and test divisions, respectively. The CIFAR-10 classes inclue images of airplanes, trucks, automobiles, ships, birds, deer, dogs, cats, frogs, and horses (Fig. 3 (c)).

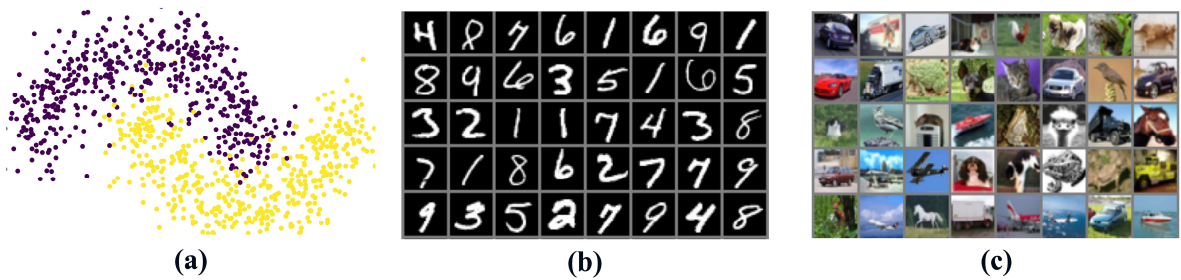

(a)          (b)          (c)

**Fig. 3.** Screenshot of evaluation datasets used for the proposed experiments including (a): Moon [64], (b): MNIST [65] and (c): CIFAR-10 [66]

### 4.1.2 Neural network architecture characteristics

In the setting of neural network architecture usage, four models with different architectures named models A, B, C, and D are used to demonstrate the generality of usage. Table 1 provides detailed information on the architectures. Model A is a simple multilayer perceptron (MLP) architecture used for toy classification problems on the Moon dataset. An Adam optimizer for model *A* with a learning rate of 0.001 is used to solve its optimization problem. Model *B* is a convolutional neural network (CNN) for the MNIST dataset, which also employs an Adam optimizer with a learning rate of 0.001. Moreover, the VGG-19 [63] as model C was used to show its wide usage for benchmark neural networks on the CIFAR-10 dataset. Finally, WideResNet [68] architecture as model *D* is used for wide CNN architectures on the CIFAR-10 dataset. Again, an Adam optimizer was employed for model *D* with a learning rate of 0.003. The proposed LSA framework could be applied to any neural network architecture with no limitations, and the aforementioned architectures A, B, C, and D in the experiments were used to provide the possibility for reproducibility of the experiments and results.

**Table 1** Experiment architectures A, B, C, and D

| Model A | $linear(100) \Rightarrow ELU \Rightarrow Linear(100) \Rightarrow ELU \Rightarrow Linear(100) \Rightarrow ELU \Rightarrow Linear(1)$ |
|---|---|
| Model B | $Conv2D(16, (5 \times 5)) \Rightarrow ReLU() \Rightarrow Conv2D(32, (5 \times 5)) \Rightarrow ReLU() \Rightarrow MaxPool2D(2,2)$ $\Rightarrow Conv2D(64, (5 \times 5)) \Rightarrow ReLU() \Rightarrow MaxPool2D(2,2) \Rightarrow Linear(100) \Rightarrow ReLU() \Rightarrow Linear(10)$ |
| Model C | VGG-19 architecture [63] |
| Model D | WideResNet [68] |

### 4.1.3   Perturbation characteristics

In the sustainability analysis scenario, the projected gradient descent (PGD) attack with $L_\infty$ norm was used. In the noisy situation of LSA's experiments, a normal Gaussian noise $N(0,1)$ and other benchmark statistical noises such as Salt, Pepper and Speckel with their default parameters are used. FGSM, PGD, TRADE, FAST, and APGD adversarial attacks were applied for different aspects of our evaluations. The characteristics of the perturbations added to the samples in each dataset are explained as follows. In the toy example model $A$ on Moon dataset, AT-FGSM was used for pure adversarial training setup, with evaluation perturbation rate set from $\varepsilon = 0.1$ up to $\varepsilon = 0.3$, the batch size set to 128, and 100 epochs for the adversarial training setup with $\varepsilon = 0.3$. Also, in experiments using model B on MNIST dataset, the evaluation perturbation rate was set from $\varepsilon = 0.1$ up to $\varepsilon = 0.3$, the batch size set at 128, and 100 epochs were passed for the adversarial training setup with $\varepsilon = 0.3$. Furthermore, in CIFAR-10 experiments on models C and D, the perturbation rate was set at $\varepsilon = 8/255$, 128 for batch size, and 150 epochs were passed for the adversarial training process. In addition, iterative white-box attacks were applied with seven iterations to evaluate the robust error in PGD evaluations with 0.005 step size.

### 4.1.4   Implementation characteristics

The implementation of the proposed framework and conduction of all experiments were carried out using PyTorch [69] library version 1.10, which allows quick manipulation of the low-level changes in neural network architectures and loss functions. Evaluations were done using NVIDIA 1080 Ti, RTX 2060 Ti, and RTX 2080 Ti graphical processors. The Scikit-learn version used was 1.02. The next section attempts to evaluate the proposed ideas and extend on experiments to evaluate different models on different datasets.

## 4.2 Experiments and results

In this section, the LSA framework (Section 3.1) is used to evaluate different network architectures with different adversarial training approaches and expose vulnerable layers. Section 4.2.1 demonstrates vulnerability feed-forward behaviors at the learnable layer level of neural networks. Section 4.2.2 instructs on the proposed methodology for decreasing layer vulnerability using the results of Section 4.2.1 as the most vulnerable layer in AT-LR adversarial training.

### 4.2.1 Evaluations of Layer Sustainability Analysis (LSA) framework

This section explains certain experiments conducted to evaluate the sustainability of each layer in a given neural network, using the Layer Sustainability Analysis (LSA) framework. The theoretical background for the LSA framework is defined in Section 3.1. As evident from Fig. 1, clean and corresponding perturbed samples are fed into each trained model. Behaviors of different model layers are first evaluated for each model architecture, using statistical perturbations followed by adversarial perturbations over the corresponding input sample. Benchmark statistical perturbations such as Gaussian, Salt, Pepper, and Speckle alongside adversarial perturbations are also used in the experiments. The identification of vulnerable layers is carried out by evaluating comparison measure (relative error) values based on Algorithm 2. The output of the LSA framework is a sorted list of the most vulnerable learnable layer numbers or an LSA MVL list. The LSA MVL list is then used in adversarial training in the next step of the AT-LR procedure, as evaluated in Section 4.2.2.

The obtained results from the LSA framework for different models, defined in Section 4.1.2, are depicted in Fig. 4. The figure illustrates comparison measure values for representation tensors of layers, during which a trained model is fed both clean and corresponding adversarially or statistically perturbed samples. Fluctuation patterns of comparison measure values for each layer in the model also demonstrate the difference in layer behaviors for clean and corresponding perturbed input. As seen in different model architectures, adversarial perturbations are more potent and have higher comparison measure values than statistical ones. In fact, as the literature shows that adversarial attacks are near the worst-case perturbations. However, the relative error of PGD-based adversarial attacks is much higher than that of FGSM adversarial attacks in all experiments. Salt and Gaussian statistical perturbation (noise) also have a much higher relative error value than the other statistical perturbations. Fig. 4.a and Algorithm 1 demonstrate that learnable layer 2 is considered a vulnerable layer for architecture model $A$ due to its higher relative error for the representation tensor of each corresponding network layer for clean and perturbed inputs, as opposed to other layers.

**Fig. 4.b**, **Fig. 4.c**, and **Fig. 4.d**, apply Algorithm 1 to architecture B, C, and D, indicating learnable layers 2, 0, and 1 as the most vulnerable learnable layers with a high relative error, respectively.

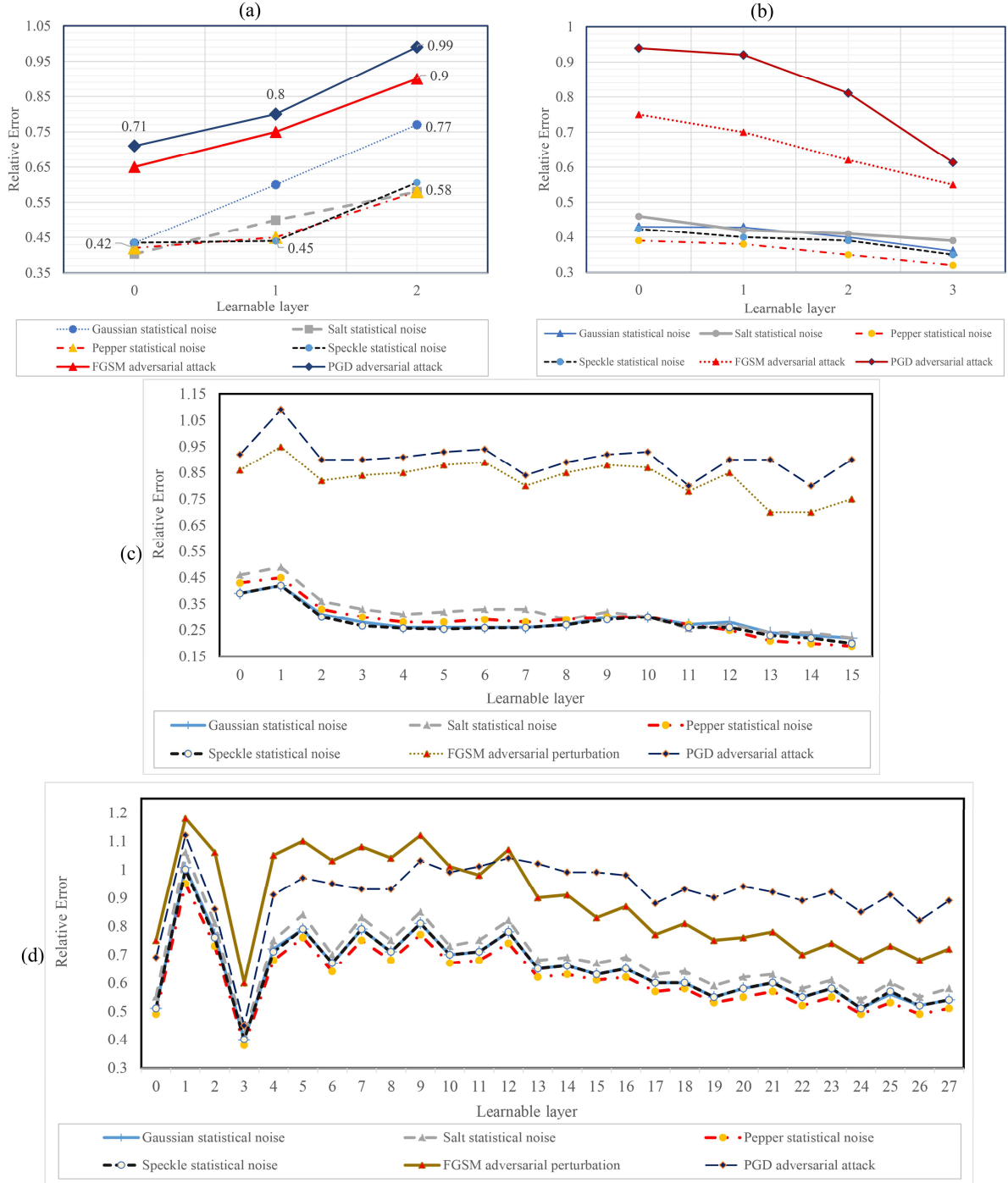

**Fig. 4.** Comparison measure values (relative error) of the LSA for statistical and adversarial perturbations on different normally trained models.

LSA results for **a)** model A, **b)** model B, **C)** model C, **d)** model D

Identifying vulnerable layers helps us to delve into interpreting behaviors of layers in the neural network models. Fig. 4 depicts such a case of inputs with corresponding perturbed versions, where the perturbation is not perceptible and does not have the identical feed-forward propagation representations in each output learnable representation layers.

### 4.2.2 Layer-wise Regularized Adversarial Training based on vulnerable layers identified by LSA

As explained in section 3.1, after using LSA to find the most vulnerable layers, proper layers were picked out as an MVL list for the LR adversarial training, named AT-LR, to achieve a more generalized and robust model. In addition, the association of each MVL proposal is indicated by appending their layer id number at the end of its model's name (e.g., AT-PGD-LR-L2 means the PGD adversarial training with the layer-wise regularized loss function for learnable layer number 2). Fig. 5 illustrates the LSA results of AT-LR adversarially trained models for corresponding models in Fig. 4. According to architecture A, learnable layer 2 has the highest comparison error and is the first item in the sorted LSA MVL list. Also, the TRADE adversarial training achieves a higher classification accuracy in much more significant perturbations than FGSM, PGD, and FAST. For architecture *A*, choosing the most vulnerable learnable layer 2 may result in a better model with the best AT-LR adversarial training approach. As illustrated in Fig. 5.a, the model AT-TRADE-LR-L2 achieves better accuracy and is able to deal with many more significant perturbations than other similar models when applied to architecture *A*. Following Fig. 4.b, the learnable layer 0 has the most significant relative error, and so, is used in AT-LR as depicted in the experiment of model B. Fig. 5.b represents the LSA framework results for normal, AT-FGSM, AT-PGD, AT-TRADE, AT-FAST, AT-APGD, along with layer-wise regularized versions such as AT-FGSM-LR-L0, AT-PGD-LR-L0, AT-TRADE-LR-L0, AT-FAST-LR-L0, AT-APGD-LR-L0. The figure demonstrates that AT-APGD-LR-L0, or the adversarially trained model through AGPD adversarial attack via regularized vulnerable learnable layer 0, could be a suitable extension on its pure adversarial training method that is more robust. The figure also exemplifies that the relative error of learnable layer 0 in AT-APGD-LR-L0 is very low, meaning that the proposed AT-LR approach has successfully enforced that layer to control its values. It is worth mentioning that near-related layers have lower values than their corresponding relative error in the standard version of the adversarial training approach. Learnable layer 0, as demonstrated, has a much lower relative error in Fig. 5.a, while AT-LR performs more efficiently in dealing with the perturbations. The LSA results for normal, adversarial training and AT-LR adversarially trained VGG-19 (model *C*), and WideResNet (model *D*) are depicted in Fig. 4.a and Fig. 4.b, respectively. Therefore, the corresponding usage of AT-LR for model *C* and *D* is done for Learnable layer 1 in both architectures, which is the vulnerable layer as depicted in Fig. 5.c and Fig. 5.d, respectively.

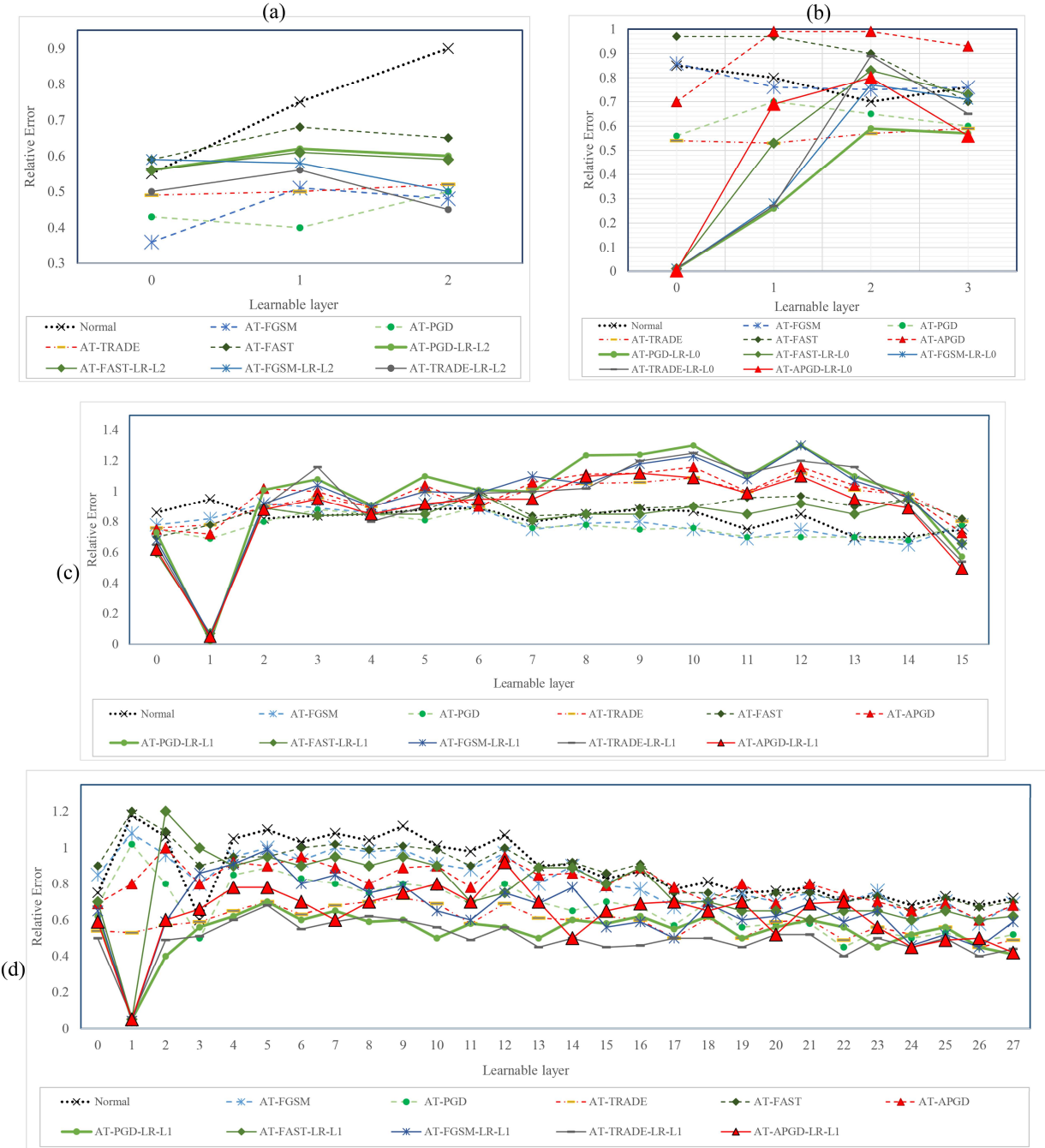

**Fig. 5.** Comparison measure values (relative error) of the LSA for statistical and adversarial perturbations on different normally trained models.

**a)** LSA results for model A, **b)** LSA results for model B, **C)** LSA results for model C, **d)** LSA results for model D

Additionally, a measure named robustness and generalization score or R&G score is defined to better evaluate each adversarially trained model over a variety of significant perturbations. Table 2 demonstrates that AT-LR reduces vulnerability, and its trained model is more robust than its normal adversarially trained one for greater perturbation rates and provides a proper generalization and robustness. In addition, a robustness and generalization score, or R&G, was defined to evaluate each adversarially trained model. The performance score value of 505.83 for AT-TRADE-LR-L1 L2 was highest among others, which is also consistent with values in Fig. 4 and Fig. 5. The top-5 values of each column are represented in bold for better readability. As can be seen, the R&G scores for AT-LR models are superior to baseline approaches. AT-FAST-LR-L2 and AT-TRADE-LR-L2 (as AT-LR models) outperform baseline adversarial training approaches by about 14.62% and 28.32% improvement in accuracy (R&G score), respectively. Furthermore, Fig. 6 illustrates the R&G score histogram of Table 2 for architecture A and represents the selection of AT-TRADE-LR-L2 as the appropriate loss function for robustifying architecture model $A$.

Table 2. Evaluation of model $A$ on Moon dataset with different loss functions against FGSM adversarial attack.

| Training type | Accuracy of model A against FGSM with different epsilon values | | | | | | |
|---|---|---|---|---|---|---|---|
| | 0 | 0.1 | 0.2 | 0.3 | 0.4 | 0.5 | R&G Score |
| Normal | **97.07** | **93.63** | 82.5 | 76.83 | 63.79 | 52.35 | 466.17 |
| AT-FGSM | **95.86** | **91.98** | **87.88** | **81.33** | 70.54 | 58.69 | **486.28** |
| AT-PGD | 86.5 | 83.89 | 80.87 | 76.61 | **72.57** | 60.02 | 460.46 |
| AT-TRADE | **96.17** | **93.06** | 85.78 | 74.2 | 68.62 | 59.68 | 477.51 |
| AT-FAST | **94.32** | 89.72 | 84.91 | 78.35 | 70.18 | 61.65 | **479.13** |
| AT-FGSM-LR-L0 | 92.81 | 90.25 | 85.36 | 80.74 | 68.21 | 58.74 | 476.11 |
| AT-FGSM-LR-L1 | 91.87 | 89.77 | **86.95** | **81.51** | 69.15 | 59.59 | 478.84 |
| AT-FGSM-LR-L2 | 93.98 | **93.01** | **89.12** | **84.01** | **75.15** | 63.25 | **498.52** |
| AT-PGD-LR-L0 | 87.02 | 84.02 | 80.64 | 76.22 | 71.12 | **65.87** | 464.89 |
| AT-PGD-LR-L1 | 86.2 | 82.94 | 79.23 | 75.16 | 70.5 | 65.04 | 459.07 |
| AT-PGD-LR-L2 | 88.78 | 86.93 | 81.25 | 77.01 | **72.81** | 66.58 | 473.36 |
| AT-FAST-LR-L0 | 86.86 | 83.85 | 80.33 | 76.34 | 71.34 | **65.79** | 464.51 |
| AT-FAST-LR-L1 | 87.91 | 85.43 | 82.44 | 77.38 | 70.01 | 62.25 | 465.42 |
| AT-FAST-LR-L2 | 92.91 | 90.31 | **86.26** | **81.53** | **75.09** | 67.65 | **493.75** |
| AT-TRADE-LR-L0 | 86.84 | 82.38 | 80.55 | 76.37 | 71.58 | 64.07 | 461.79 |
| AT-TRADE-LR-L1 | 87.25 | 84.22 | 81.25 | 76.98 | 71.14 | 63.25 | 464.09 |
| AT-TRADE-LR-L2 | **96.67** | **93.2** | **86.56** | **81.25** | **79.5** | **68.65** | **505.83** |

It is important to note that the attack methodology is effective for adversarial training. Further, using a new state-of-the-art benchmark adversarial training approach with our AT-LR scenario might be appropriate for reducing layer vulnerabilities through LSA outputs. The results of experiments for models B, C, and D are presented in Table 3 and Table 4. As evident, the AT-APGD-LR-L1 model on architectures C and D outperforms other related models in CIFAR-10 with improvements in the magnitudes of 8.52% and 7.55%, respectively. More results of different models are available along with codesheets on Github at https://github.com/khalooei/LSA.

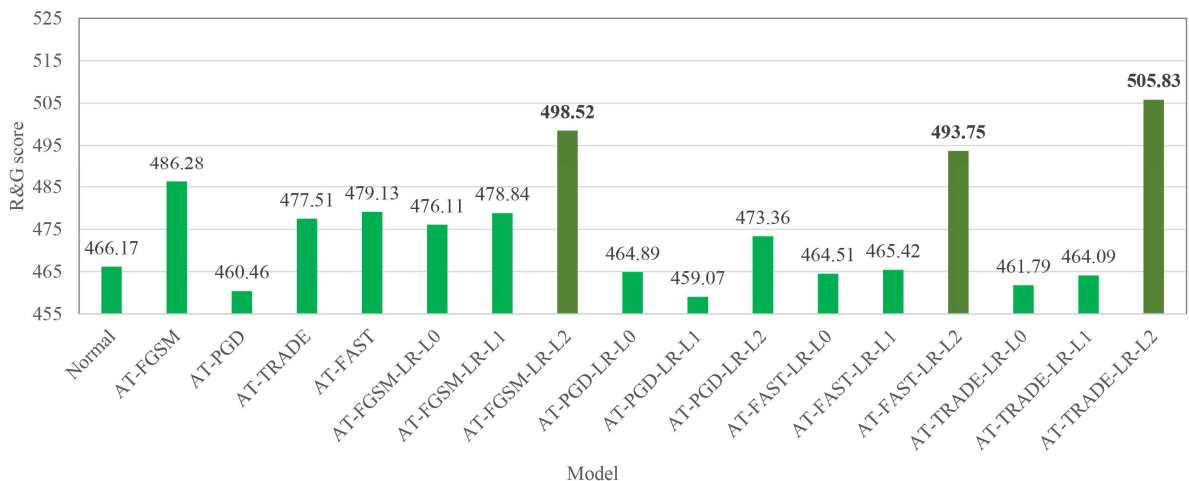

**Fig. 6.** Comparison of R&G score results for different loss functions of architecture A on Moon dataset with different training loss functions.

**Table 3**. Evaluation of model *B* on MNIST dataset with different loss functions against FGSM adversarial attack.

| Training type | Accuracy of model A against FGSM with different epsilon values | | | | | | |
|---|---|---|---|---|---|---|---|
| | 0 | 0.1 | 0.2 | 0.3 | 0.4 | 0.5 | R&G Score |
| Normal | 98.82 | 82.1 | 47.2 | 17.87 | 6.96 | 4.25 | 257.2 |
| AT-APGD | **99.36** | **98.54** | **97.67** | 96.97 | 71.72 | 33.73 | 497.99 |
| AT-APGD-LR-L0 | 98.97 | 98.11 | 97.36 | **97.17** | **94.6** | **55.52** | **541.73** |

**Table 4**. Evaluation of models *C* and *D* on CIFAR-10 with different loss functions against FGSM adversarial attack.

| Architecture | Training type | Accuracy of model C and D against FGSM with different epsilon values | | | | | |
|---|---|---|---|---|---|---|---|
| | | 0 | 0.01 | 0.03 | 0.1 | 0.2 | R&G Score |
| Model C | Normal | **90.53** | 48.91 | 43.91 | 31.5 | 22.29 | 237.14 |
| | AT-APGD | 83.68 | 74.69 | 72.21 | 50.04 | 41.45 | 322.07 |
| | AT-APGD-LR-L1 | 83.55 | **75.32** | **73.35** | **52.36** | **46.01** | **330.59** |
| Model D | Normal | **90.01** | 40.01 | 36.35 | 22.3 | 16.2 | 204.87 |
| | AT-APGD | 82.60 | 50.12 | **47.11** | 41.46 | 40.01 | 261.3 |
| | AT-APGD-LR-L1 | 81.81 | **53.21** | 48.17 | **43.01** | **42.65** | **268.85** |

The following section discusses relevant analysis of decision boundaries as an intriguing element in the proposed AT-LR method of this paper. The effectiveness of each proposal item of the LSA list for AT-LR adversarial trained models leading to different decision boundaries are illustrated, along with intriguing properties of AT-LR adversarial training on the loss functions, which are valuable in explaining and interpreting the behaviors of neural networks.

### 4.2.3 Intriguing behaviors of decision boundary on AT-LR approach for each LSA MVL proposal

A toy example that illustrates the different loss landscapes in a two-dimensional view of the decision boundary can be used to evaluate the different aspects of the proposed AT-LR adversarial training loss function based on model A and the 2D Moon dataset. All of the configurations and settings are based on those mentioned in Section 4.1. Also, a specific seed value is used for any randomization functions to attain better reproducibility of results. Fig. 7 shows the TRADE adversarial training (AT-TRADE) decision boundary and other AT-LR trained models based on their learnable layer numbers of all LSA MVL items (further illustrations and results are available at https://github.com/khalooei/LSA). The figure also shows estimated decision boundaries and loss landscapes corresponding to the various loss functions used in the training phase, with an individual sample (shown in yellow) indicated for each decision boundary diagram, as well as its corresponding adversarial sample as shown in red (with FGSM adversarial attack and perturbation rate of 0.3 in $l_\infty$ bound). Instead of cluttering the decision boundary with clean samples and in order to better visualize 2D presentations, 1000 adversarial sample points are labeled as adversarial points with their corresponding accurate color-codes (brown and green points represent sample points from different classes). Depending on the loss function curvature of the model, different models behave differently and produce different adversarial examples for a given sample point. This demonstrates that using AT-LR for more vulnerable layers identified by the LSA framework ameliorates the decision boundaries. An example of such amelioration can be seen in Fig. 7.d for decision boundaries of vulnerable layer 2 in model A on the Moon dataset (AT-TRAED-LR-L2).

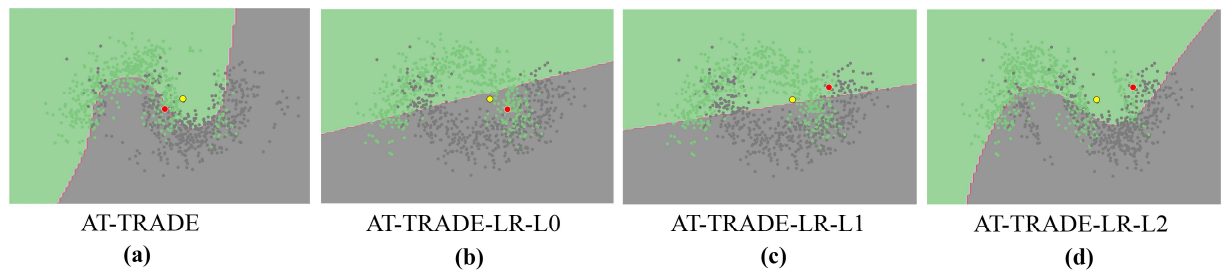

| AT-TRADE | AT-TRADE-LR-L0 | AT-TRADE-LR-L1 | AT-TRADE-LR-L2 |
| (a) | (b) | (c) | (d) |

**Fig. 7.** Decision boundary of adversarial training with different loss functions on Model A.

# 5  Conclusion and future work

This paper proposed an efficient Layer Sustainability Analysis (LSA) framework based on the Lipschitz theory. Any DNN can take advantage of our proposed framework. Finding the Most Vulnerable Layers (MVL) in a DNN (aka LSA MVL list or MVL proposals) can provide insights into procuring a remedy for different perturbations. To this end, both clean and corresponding perturbed samples were fed to the DNN. For each sample, the layer output representation tensors corresponding to the clean and corresponding perturbed samples were compared and the relative difference between the two representations was used to calculate the level of vulnerability. The investigated layers were exploited in a Layer-wise Regularized Adversarial Training (aka AT-LR) scenario to improve the conventional adversarial training method. As well as assessing general sustainability in the face of a wide array of perturbations, the LSA also provides a better understanding of behavioral outcomes and capabilities. The AT-LR could be used with any benchmark adversarial attack approaches in the LSA framework to reduce the vulnerability of each network layer. Also, the experiments are made a bit fair and valuable by defining a regularization and generalization score called the R&G score. On the Moon and MNIST datasets, AT-TRADE-LR-L2 and AT-APGD-LR-L0 models outperform the other related models with 28.32% and 43.74% R&G score improvement, respectively. In addition, the AT-APGD-LR-L1 model on architectures VGG-19 and WideResNet outperforms the other related models in CIFAR-10 with improvements of 8.52% and 7.55%, respectively. The proposed idea performs well both theoretically and experimentally on benchmark Moon, MNIST, and CIFAR-10 datasets comparable to most of the published works such as AT-TRADE, AT-FAST, and AT-APGD. Incorporating this approach with other related state-of-the-art adversarial defense techniques is a step towards improving robustness and generalization. Future works are encouraged to use different comparison measures with an assessment of the output representation tensor of layers. Alternatively, metric learning approaches or a distributionally robust optimization problem could be applied to the comparison measure as part of the LSA framework. Also, a layer-level interpretation could be improved by using and integrating attention mechanisms in the LSA framework. It is worth noting that any reductions in vulnerability of items in the LSA MVL list can act as a good alternative for AT-LR. The LSA framework code and documents are available and open-sourced at https://github.com/khalooei/LSA.

**Acknowledgments**

This publication was supported by grant No. RD-51-9911-0025 from the R&D Center of Mobile Telecommunication Company of Iran (MCI) for advancing information and communications technologies. Additionally, we are grateful to the computer science department of the Institute for Research in Fundamental Sciences (IPM) for the provision of a part of our needs to graphical processing servers for experimental purposes.

**References**

[1]  M. Alam, M.D. Samad, L. Vidyaratne, A. Glandon, K.M. Iftekharuddin, Survey on Deep Neural Networks in Speech and Vision Systems, Neurocomputing. 417 (2020) 302–321.

[2]  B. Lim, S. Zohren, Time-series forecasting with deep learning: a survey, Philosophical Transactions of the Royal Society A. 379 (2021).

[3]  F. Piccialli, V. Di Somma, F. Giampaolo, S. Cuomo, G. Fortino, A survey on deep learning in medicine: Why, how and when?, Information Fusion. 66 (2021) 111–137.

[4]  W. Liu, Z. Wang, X. Liu, N. Zeng, Y. Liu, F.E. Alsaadi, A survey of deep neural network architectures and their applications, Neurocomputing. 234 (2017) 11–26.

[5]  C. Szegedy, W. Zaremba, I. Sutskever, J. Bruna, D. Erhan, I. Goodfellow, R. Fergus, Intriguing properties of neural networks, in: International Conference on Learning Representations, International Conference on Learning Representations (ICLR), 2014.

[6]  I.J. Goodfellow, J. Shlens, C. Szegedy, Explaining and Harnessing Adversarial Examples, in: International Conference on Learning Representations, 2015.

[7]  N. Akhtar, A. Mian, Threat of Adversarial Attacks on Deep Learning in Computer Vision: A Survey, IEEE Access. 6 (2018) 14410–14430.

[8]  A. Madry, A. Makelov, L. Schmidt, D. Tsipras, A. Vladu, Towards Deep Learning Models Resistant to Adversarial Attacks, in: International Conference on Learning Representations, 2018.

[9]  C. Xiao, J.-Y. Zhu, B. Li, W. He, M. Liu, D. Song, Spatially Transformed Adversarial Examples, (2018). http://arxiv.org/abs/1801.02612 (accessed August 18, 2019).

[10]  S. Gu, L. Rigazio, Towards deep neural network architectures robust to adversarial examples, in: International Conference on Learning Representations Workshop, International Conference on Learning Representations, ICLR, 2015.

[11]  Y. Jang, T. Zhao, S. Hong, H. Lee, Adversarial Defense via Learning to Generate Diverse Attacks, in: IEEE International Conference on Computer Vision, 2019.

[12]  L. Schmidt, S. Santurkar, D. Tsipras, K. Talwar, A. Madry, Adversarially Robust Generalization Requires More Data, in: S. Bengio, H. Wallach, H. Larochelle, K. Grauman, N. Cesa-Bianchi, R. Garnett (Eds.), International Conference on Neural Information Processing Systems, Curran Associates, Inc., 2018: pp. 5019–5031.

[13]  C. Xie, Y. Wu, L. van der Maaten, A.L. Yuille, K. He, Feature Denoising for Improving Adversarial Robustness, in: IEEE Conference on Computer Vision and Pattern Recognition, 2019.

[14]  F. Liao, M. Liang, Y. Dong, T. Pang, X. Hu, J. Zhu, Defense Against Adversarial Attacks Using High-Level Representation Guided Denoiser, in: IEEE Conference on Computer Vision and Pattern Recognition, 2018.

[15]  W. Xu, D. Evans, Y. Qi, Feature Squeezing: Detecting Adversarial Examples in Deep Neural Networks, Network and Distributed Systems Security Symposium (NDSS) 2018. (2018).

[16]  H.M. Arjomandi, M. Khalooei, M. Amirmazlaghani, Limited Budget Adversarial Attack Against Online Image Stream, in: International Conference on Machine Learning Workshop on Adversarial Machine Learning, 2021.

[17]  H. Zhang, M. Cisse, Y.N. Dauphin, D. Lopez-Paz, mixup: Beyond empirical risk minimization, in: International Conference on Learning Representations, 2018.

[18]  D. Stutz, M. Hein, B. Schiele, Disentangling Adversarial Robustness and Generalization, in: IEEE Conference on Computer Vision and Pattern Recognition, 2019.

[19]  W. Wei, L. Liu, M. Loper, K.H. Chow, E. Gursoy, S. Truex, Y. Wu, Cross-Layer Strategic Ensemble Defense Against Adversarial Examples, in: International Conference on Computing, Networking and Communications, Institute of Electrical and Electronics Engineers Inc., 2020: pp. 456–460.

[20]  A. Sinha, H. Namkoong, J. Duchi, Certifying Some Distributional Robustness with Principled Adversarial

Training, in: International Conference on Learning Representations, 2018.

[21]   F. Tramèr, A. Kurakin, N. Papernot, I. Goodfellow, D. Boneh, P. McDaniel, Ensemble Adversarial Training: Attacks and Defenses, International Conference on Learning Representations. (2018).

[22]   N. Carlini, A. Athalye, N. Papernot, W. Brendel, J. Rauber, D. Tsipras, I. Goodfellow, A. Madry, A. Kurakin, On Evaluating Adversarial Robustness, ArXiv Preprint ArXiv:1902.06705. (2019).

[23]   S. Sabour, Y. Cao, F. Faghri, D.J. Fleet, Adversarial Manipulation of Deep Representations, in: International Conferenceon Learning Representations, 2016.

[24]   S. Sankaranarayanan, A. Jain, R. Chellappa, S.N. Lim, Regularizing Deep Networks Using Efficient Layerwise Adversarial Training, AAAI Conference on Artificial Intelligence. 32 (2018).

[25]   X. Chen, N. Zhang, Layer-wise Adversarial Training Approach to Improve Adversarial Robustness, International Joint Conference on Neural Networks. (2020).

[26]   A. Kurakin, I. Goodfellow, S. Bengio, Adversarial examples in the physical world, in: International Conference on Learning Representations Workshop, 2016.

[27]   F. Croce, M. Hein, Reliable evaluation of adversarial robustness with an ensemble of diverse parameter-free attacks, in: H.D. III, A. Singh (Eds.), International Conference on Machine Learning, PMLR, 2020: pp. 2206–2216.

[28]   S. Bubeck, Y.T. Lee, E. Price, I. Razenshteyn, Adversarial examples from computational constraints, in: K. Chaudhuri, R. Salakhutdinov (Eds.), International Conference on Machine Learning, PMLR, 2019: pp. 831–840.

[29]   M. Paknezhad, C.P. Ngo, A.A. Winarto, A. Cheong, B.C. Yang, W. Jiayang, L.H. Kuan, Explaining adversarial vulnerability with a data sparsity hypothesis, Neurocomputing. (2022).

[30]   A. Athalye, L. Engstrom, A. Ilyas, K. Kwok, Synthesizing Robust Adversarial Examples, in: J. Dy, A. Krause (Eds.), International Conference on Machine Learning, PMLR, 2018: pp. 284–293.

[31]   S. Zheng, Y. Song, T. Leung, I. Goodfellow, Improving the Robustness of Deep Neural Networks via Stability Training, IEEE Conference on Computer Vision and Pattern Recognition. (2016) 4480–4488.

[32]   H. Zhang, Y. Yu, J. Jiao, E.P. Xing, L. El Ghaoui, M.I. Jordan, Theoretically Principled Trade-off between

Robustness and Accuracy, International Conference on Machine Learning. (2019) 12907–12929.

[33] A. Kurakin, I. Goodfellow, S. Bengio, Y. Dong, F. Liao, M. Liang, T. Pang, J. Zhu, X. Hu, C. Xie, J. Wang, Z. Zhang, Z. Ren, A. Yuille, S. Huang, Y. Zhao, Y. Zhao, Z. Han, J. Long, Y. Berdibekov, T. Akiba, S. Tokui, M. Abe, Adversarial Attacks and Defences Competition, ArXiv. abs/1804.0 (2018).

[34] T. Pang, K. Xu, C. Du, N. Chen, J. Zhu, Improving Adversarial Robustness via Promoting Ensemble Diversity, in: K. Chaudhuri, R. Salakhutdinov (Eds.), International Conference on Machine Learning, PMLR, 2019: pp. 4970–4979.

[35] T. Pang, K. Xu, Y. Dong, C. Du, N. Chen, J. Zhu, Rethinking Softmax Cross-Entropy Loss for Adversarial Robustness, in: International Conference on Learning Representations, 2020.

[36] J. Zhang, X. Xu, B. Han, G. Niu, L. Cui, M. Sugiyama, M.S. Kankanhalli, Attacks Which Do Not Kill Training Make Adversarial Learning Stronger, in: International Conference on Machine Learning, 2020: pp. 11278–11287.

[37] A. Raghunathan, J. Steinhardt, P. Liang, Certified Defenses against Adversarial Examples, in: International Conference on Learning Representations, 2018.

[38] A. Chan, Y. Tay, Y. Ong, J. Fu, Jacobian Adversarially Regularized Networks for Robustness, in: International Conference on Learning Representations, 2020.

[39] A. ArjomandBigdeli, M. Amirmazlaghani, M. Khalooei, Defense against adversarial attacks using DRAGAN, in: Iranian Conference on Signal Processing and Intelligent Systems, 2020: pp. 1–5.

[40] P. Xia, B. Li, Improving resistance to adversarial deformations by regularizing gradients, Neurocomputing. 455 (2021) 38–46.

[41] E. Raff, J. Sylvester, S. Forsyth, M. McLean, Barrage of Random Transforms for Adversarially Robust Defense, in: IEEE Conference on Computer Vision and Pattern Recognition, 2019: pp. 6521–6530.

[42] C. Xie, J. Wang, Z. Zhang, Z. Ren, A. Yuille, Mitigating adversarial effects through randomization, in: International Conference on Learning Representations, 2018.

[43] C. Guo, M. Rana, M. Cissé, L.V.D. Maaten, Countering Adversarial Images using Input Transformations, in: International Conference on Learning Representations, 2018.

[44]   P. Samangouei, M. Kabkab, R. Chellappa, Defense-GAN: Protecting Classifiers Against Adversarial Attacks Using Generative Models, in: International Conference on Learning Representations, 2018.

[45]   A. Mustafa, S. Khan, M. Hayat, R. Goecke, J. Shen, L. Shao, Adversarial Defense by Restricting the Hidden Space of Deep Neural Networks, in: IEEE International Conference on Computer Vision, 2019.

[46]   R. Sahay, R. Mahfuz, A.E. Gamal, Combatting Adversarial Attacks through Denoising and Dimensionality Reduction: A Cascaded Autoencoder Approach, in: Annual Conference on Information Sciences and Systems (CISS), 2019: pp. 1–6.

[47]   D. Meng, H. Chen, Magnet: a two-pronged defense against adversarial examples, in: ACM Conference on Computer and Communications Security, ACM Press, New York, New York, USA, 2017: pp. 135–147.

[48]   Y. Wang, W. Zhang, T. Shen, H. Yu, F.Y. Wang, Binary thresholding defense against adversarial attacks, Neurocomputing. 445 (2021) 61–71.

[49]   S. lin Yin, X. lan Zhang, L. yu Zuo, Defending against adversarial attacks using spherical sampling-based variational auto-encoder, Neurocomputing. 478 (2022) 1–10.

[50]   Q. Guo, J. Ye, Y. Chen, Y. Hu, Y. Lan, G. Zhang, X. Li, INOR—An Intelligent noise reduction method to defend against adversarial audio examples, Neurocomputing. 401 (2020) 160–172.

[51]   F. Crecchi, M. Melis, A. Sotgiu, D. Bacciu, B. Biggio, FADER: Fast adversarial example rejection, Neurocomputing. 470 (2022) 257–268.

[52]   A. Fawzi, H. Fawzi, O. Fawzi, Adversarial Vulnerability for Any Classifier, in: International Conference on Neural Information Processing Systems, Curran Associates Inc., Red Hook, NY, USA, 2018: pp. 1186–1195.

[53]   E. Wong, Z. Kolter, Provable Defenses against Adversarial Examples via the Convex Outer Adversarial Polytope, in: International Conference on Machine Learning, 2018: pp. 5286–5295.

[54]   E. Wong, F. Schmidt, J.H. Metzen, J.Z. Kolter, Scaling provable adversarial defenses, in: S. Bengio, H. Wallach, H. Larochelle, K. Grauman, N. Cesa-Bianchi, R. Garnett (Eds.), International Conference on Neural Information Processing Systems, Curran Associates, Inc., 2018: pp. 8410–8419.

[55]   A. Raghunathan, J. Steinhardt, P.S. Liang, Semidefinite relaxations for certifying robustness to adversarial

examples, in: S. Bengio, H. Wallach, H. Larochelle, K. Grauman, N. Cesa-Bianchi, R. Garnett (Eds.), International Conference on Neural Information Processing Systems, Curran Associates, Inc., 2018.

[56]   N. Carlini, G. Katz, C. Barrett, D. Dill, Provably Minimally-Distorted Adversarial Examples, ArXiv: Learning. (2017).

[57]   D. Tsipras, S. Santurkar, L. Engstrom, A. Turner, A. Madry, Robustness May Be at Odds with Accuracy, International Conference on Learning Representations. (2018).

[58]   L. Rice, E. Wong, Z. Kolter, Overfitting in adversarially robust deep learning, in: H.D. III, A. Singh (Eds.), International Conference on Machine Learning, PMLR, 2020: pp. 8093–8104.

[59]   J. Zhang, J. Zhu, G. Niu, B. Han, M. Sugiyama, M. Kankanhalli, Geometry-aware Instance-reweighted Adversarial Training, in: International Conference on Learning Representations, 2021. https://openreview.net/forum?id=iAX0l6Cz8ub.

[60]   T. Chen, Z. Zhang, S. Liu, S. Chang, Z. Wang, T. Ima-, Robust Overfitting May Be Mitigated By Properly -, in: International Conference on Learning Representations, 2021: pp. 1–19.

[61]   E. Wong, L. Rice, J.Z. Kolter, Fast is better than free: Revisiting adversarial training, in: International Conference on Learning Representations, 2020.

[62]   M. Andriushchenko, N. Flammarion, Understanding and Improving Fast Adversarial Training, in: International Conference on Neural Information Processing Systems, 2020: pp. 16048–16059.

[63]   K. Simonyan, A. Zisserman, Very deep convolutional networks for large-scale image recognition, in: International Conference on Learning Representations, 2015.

[64]   sklearn.datasets.make_moons — scikit-learn 1.0 documentation, (n.d.). https://scikit-learn.org/stable/modules/generated/sklearn.datasets.make_moons.html (accessed September 28, 2021).

[65]   C.C. and C.B. Yann LeCun, MNIST handwritten digit database, (n.d.). http://yann.lecun.com/exdb/mnist/ (accessed June 24, 2019).

[66]   G.H. Alex Krizhevsky, Vinod Nair, CIFAR-10 and CIFAR-100 datasets, (2009). https://www.cs.toronto.edu/~kriz/cifar.html (accessed October 19, 2019).

[67]   F. Pedregosa, G. Varoquaux, A. Gramfort, V. Michel, B. Thirion, O. Grisel, M. Blondel, P. Prettenhofer, R.

Weiss, V. Dubourg, J. Vanderplas, A. Passos, D. Cournapeau, M. Brucher, M. Perrot, É. Duchesnay, Scikit-learn: Machine Learning in Python, Journal of Machine Learning Research. 12 (2011) 2825–2830.

[68] S. Zagoruyko, N. Komodakis, Wide Residual Networks, in: British Machine Vision Conference, British Machine Vision Association, 2016.

[69] A. Paszke, S. Gross, F. Massa, A. Lerer, J. Bradbury, G. Chanan, T. Killeen, Z. Lin, N. Gimelshein, L. Antiga, A. Desmaison, A. Köpf, E. Yang, Z. DeVito, M. Raison, A. Tejani, S. Chilamkurthy, B. Steiner, L. Fang, J. Bai, S. Chintala, PyTorch: An Imperative Style, High-Performance Deep Learning Library, in: International Conference on Neural Information Processing Systems, 2019.

# Appendix

## A. Model Architectures

**Table 1** Model A, B, C and D architectures

| Model | Architecture |
|---|---|
| Model A | $linear(100) \Rightarrow ELU \Rightarrow Linear(100) \Rightarrow ELU \Rightarrow Linear(200) \Rightarrow ELU \Rightarrow Linear(1)$ |
| Model B | $Conv2D(16, (5 \times 5)) \Rightarrow ReLU() \Rightarrow Conv2D(32, (5 \times 5)) \Rightarrow ReLU()$ $\Rightarrow MaxPool2D(2,2)$ $\Rightarrow Conv2D(64, (5 \times 5)) \Rightarrow ReLU() \Rightarrow MaxPool2D(2,2) \Rightarrow Linear(100) \Rightarrow ReLU$ $\Rightarrow Linear(10)$ |
| Model C | $Conv2D(64, (3 \times 3)) \Rightarrow BatchNorm2D() \Rightarrow ReLU()$ $\Rightarrow Conv2D(64, (3 \times 3)) \Rightarrow BatchNorm2D() \Rightarrow ReLU()$ $\Rightarrow MaxPool2D(2,2) \Rightarrow Conv2D(128, (3 \times 3)) \Rightarrow BatchNorm2D() \Rightarrow ReLU()$ $\Rightarrow MaxPool2D(2,2) \Rightarrow Conv2D(256, (3 \times 3)) \Rightarrow BatchNorm2D() \Rightarrow ReLU()$ $\Rightarrow MaxPool2D(2,2) \Rightarrow Conv2D(256, (3 \times 3)) \Rightarrow BatchNorm2D() \Rightarrow ReLU()$ $\Rightarrow Conv2D(512, (3 \times 3)) \Rightarrow BatchNorm2D() \Rightarrow ReLU()$ $\Rightarrow Conv2D(512, (3 \times 3)) \Rightarrow BatchNorm2D() \Rightarrow ReLU()$ $\Rightarrow Conv2D(512, (3 \times 3)) \Rightarrow BatchNorm2D() \Rightarrow ReLU()$ $\Rightarrow Conv2D(512, (3 \times 3)) \Rightarrow BatchNorm2D() \Rightarrow ReLU()$ $\Rightarrow Conv2D(512, (3 \times 3)) \Rightarrow BatchNorm2D() \Rightarrow ReLU()$ $\Rightarrow Conv2D(512, (3 \times 3)) \Rightarrow BatchNorm2D() \Rightarrow ReLU()$ $\Rightarrow Conv2D(512, (3 \times 3)) \Rightarrow BatchNorm2D() \Rightarrow ReLU()$ $\Rightarrow MaxPool2D(2,2) \Rightarrow AvgPool2D(1,1) \Rightarrow Linear(10)$ |
| Model D (WideResNet [54]) | $Conv2D(16, (3 \times 3)) \Rightarrow BatchNorm2D() \Rightarrow LeakyReLU() \Rightarrow Conv2D(160, (3 \times 3))$ $\Rightarrow$ $Block1: [BatchNorm2D() \Rightarrow LeakyReLU() \Rightarrow Conv2D(160, (3 \times 3))$ $\Rightarrow AvgPool2D(1,1) \Rightarrow ChannelPadding()] \Rightarrow$ $Block2: [BatchNorm2D() \Rightarrow LeakyReLU() \Rightarrow Conv2D(320, (3 \times 3))$ $\Rightarrow AvgPool2D(2,2) \Rightarrow ChannelPadding()] \Rightarrow$ $Block3: [BatchNorm2D() \Rightarrow LeakyReLU() \Rightarrow Conv2D(640, (3 \times 3))$ $\Rightarrow AvgPool2D(2,2) \Rightarrow ChannelPadding()] \Rightarrow$ $BatchNorm2D() \Rightarrow LeakyReLU() \Rightarrow linear(10)$ |